\documentclass[sigconf]{aamas}

\usepackage{balance} 
\usepackage{mathtools}
\usepackage{algorithm}
\usepackage{algorithmic}

\makeatletter
\gdef\@copyrightpermission{
  \begin{minipage}{0.2\columnwidth}
   \href{https://creativecommons.org/licenses/by/4.0/}{\includegraphics[width=0.90\textwidth]{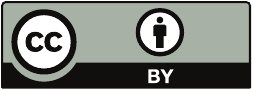}}
  \end{minipage}\hfill
  \begin{minipage}{0.8\columnwidth}
   \href{https://creativecommons.org/licenses/by/4.0/}{This work is licensed under a Creative Commons Attribution International 4.0 License.}
  \end{minipage}
  \vspace{5pt}
}
\makeatother

\setcopyright{ifaamas}
\acmConference[AAMAS '25]{Proc.\@ of the 24th International Conference
on Autonomous Agents and Multiagent Systems (AAMAS 2025)}{May 19 -- 23, 2025}
{Detroit, Michigan, USA}{Y.~Vorobeychik, S.~Das, A.~Nowé  (eds.)}
\copyrightyear{2025}
\acmYear{2025}
\acmDOI{}
\acmPrice{}
\acmISBN{}

\title[AAMAS-2025 Formatting Instructions]{Evaluation-Time Policy Switching for Offline Reinforcement Learning}
\subtitle{AAAI Track}

\author{Natinael Solomon Neggatu}
\affiliation{
  \institution{University of Warwick}
  \city{Coventry}
  \country{United Kingdom}}
\email{nati.neggatu@warwick.ac.uk}

\author{Jeremie Houssineau}
\affiliation{
  \institution{Nanyang Technological University}
  \city{}
  \country{Singapore}}
\email{jeremie.houssineau@ntu.edu.sg}

\author{Giovanni Montana}
\affiliation{
  \institution{University of Warwick}
  \city{Coventry}
  \country{United Kingdom}}
\email{g.montana@warwick.ac.uk}

\begin{abstract}
Offline reinforcement learning (RL) looks at learning how to optimally solve tasks using a fixed dataset of interactions from the environment. Many off-policy algorithms developed for online learning struggle in the offline setting as they tend to over-estimate the behaviour of out of distributions actions. Existing offline RL algorithms adapt off-policy algorithms, employing techniques such as constraining the policy or modifying the value function to achieve good performance on individual datasets but struggle to adapt to different tasks or datasets of different qualities without tuning hyper-parameters. We introduce a policy switching technique that dynamically combines the behaviour of a pure off-policy RL agent, for improving behaviour, and a behavioural cloning (BC) agent, for staying close to the data. We achieve this by using a combination of epistemic uncertainty, quantified by our RL model, and a metric for aleatoric uncertainty extracted from the dataset. We show empirically that our policy switching technique can outperform not only  the individual algorithms used in the switching process but also compete with state-of-the-art methods on numerous benchmarks. Our use of epistemic uncertainty for policy switching also allows us to naturally extend our method to the domain of offline to online fine-tuning allowing our model to adapt quickly and safely from online data, either matching or exceeding the performance of current methods that typically require additional  modification or hyper-parameter fine-tuning.
\end{abstract}


\newcommand{\BibTeX}{\rm B\kern-.05em{\sc i\kern-.025em b}\kern-.08em\TeX}

\begin{document}


\pagestyle{fancy}
\fancyhead{}


\maketitle 


\section{Introduction}

Reinforcement learning (RL) is a powerful framework for solving sequential decision-making problems. Recent advancements integrating deep learning have shown RL's success across various domains, e.g., game playing \citep{silver2017mastering, mnih2015human}, recommendation systems \citep{zou2019reinforcement}, and protein folding \citep{jumper2021highly}. While online RL involves an agent interacting directly with an environment to learn optimal behaviour, offline RL focuses on extracting effective policies from fixed datasets without environmental interaction. This approach is particularly valuable in domains where exploration or data collection can be costly or dangerous, such as robotics \citep{levine2018learning, chebotar2021actionable}, healthcare \citep{nambiar2023deep, tang2022leveraging}, and autonomous driving \citep{shi2021offline, li2023boosting}. 

A primary challenge in offline RL is safely generalising behaviour from limited datasets. With finite data, agents often poorly estimate the value of out-of-distribution (OOD) state-action pairs, leading to the exploitation of overestimated OOD actions. This exploitation typically results in distributional shift and performance degradation. To mitigate these issues, many offline RL algorithms adapt off-policy methods to constrain learned policies to stay close to the available data. These approaches generally fall into two categories: policy constraint methods, which explicitly penalise behaviours that deviate from the dataset \citep{kumar2019stabilizing, nair2020awac, brandfonbrener2021offline, wu2019behavior, fujimoto2021minimalist}, and value regularisation methods, which implicitly affect agent behaviour by modifying the value function for OOD actions \citep{an2021uncertainty, kumar2020conservative, bai2022pessimistic}. 

While these approaches have shown promise, significant challenges remain in offline RL. Current methods often struggle to balance the exploitation of valuable information in the dataset with the need to generalise safely to unseen situations. This balance is crucial for achieving good performance across a range of tasks and dataset qualities. Instead, existing algorithms frequently require extensive hyper-parameter tuning for individual tasks and datasets, limiting their practical applicability. Furthermore, as shown by \citet{nair2020awac} and \citet{lee2022offline}, the constraints imposed during offline learning can hinder adaptation during online fine-tuning, making it difficult for agents to effectively incorporate new experiences.

In this work, we explore a novel approach to address these challenges by leveraging the complementary strengths of reinforcement learning and behavioural cloning (BC). We investigate the potential of a policy switching method that dynamically combines these two paradigms. This approach aims to harness the RL agent's ability to construct effective trajectories using dynamic programming techniques, while utilising BC to provide actions closely aligned with the dataset when necessary. Unlike previous methods that incorporate supervised learning as a regularisation term during training, our approach maintains a clear separation between RL and BC. This allows for independent training of each element, preserving the integrity of their respective learning processes. The core of our method lies in the switching process, which leverages both epistemic uncertainty (arising from the learning process) and aleatoric uncertainty (inherent in the dataset) to guide the dynamic selection of actions.

A distinctive aspect of our method is its implementation at evaluation time. Unlike most offline RL techniques that modify the training process itself, our approach leaves the training of the RL and BC policies unchanged. Instead, we introduce the policy switching mechanism only when the trained policies are being used to make decisions. This evaluation-time approach aims to provide greater flexibility and adaptability across different tasks and dataset qualities without requiring extensive retraining or hyper-parameter tuning of the underlying policies.

We investigate the effects of this novel approach on controlling agent behaviour in unobserved environmental regions using various offline RL benchmarks. To our knowledge, this is the first study exploring policy switching in offline RL algorithms at evaluation time, rather than during training, to improve offline performance. Our study explores whether operating at evaluation time can improve performance by adapting to specific situations without altering the core training process. We also examine the potential of this approach to reduce requirements for, and facilitate more effective online fine-tuning, potentially allowing for quick adaptation to new data without compromising the original learned policies.
Through our experiments, we show that our policy switching method can lead to better performance both in offline settings and during the transition from offline to online tuning, without the need for extensive dataset-specific tuning that is common in existing approaches.

\section{Related work}

\subsection{Offline RL}

Offline RL methods typically balance improving behaviour using dynamic programming (DP) principles \citep{bellman1966dynamic} with staying close to the available data. This balance is achieved through two main approaches. Policy constraint methods either explicitly \citep{fujimoto2019off, fujimoto2021minimalist} or implicitly \citep{nair2020awac, kumar2019stabilizing, zhou2021plas, wu2019behavior} enforce the agent to stay close to the data. However, the performance of these methods can be susceptible to the quality of the policy generating the dataset.

Conservative methods, alternatively, pessimistically estimate the expected return of out-of-distribution (OOD) actions \citep{an2021uncertainty, kumar2020conservative, bai2022pessimistic}. CQL \citep{kumar2020conservative} achieves this by uniformly penalising the value of OOD actions generated by the learned policy. SAC-N \citep{an2021uncertainty} modifies the SAC \citep{haarnoja2018soft} algorithm to include a variable number of critics, using the variance across the ensemble to penalise value estimates. This approach dynamically adjusts the degree of pessimism based on how far actions are from the data support, mitigating the issue of overestimation bias for OOD actions \citep{van2016deep}. As a result, they are able to achieve SOTA performance with substantially fewer hyper-parameters compared to methods such as CQL and RORL \citep{yang2022rorl}.

Despite the benefits of using uncertainty to guide learning, the methods proposed by \citet{an2021uncertainty} are highly sensitive to ensemble size, with no clear method to infer the appropriate size based on the task or data quality. TD3-BC-N \citep{beeson2024balancing} address this by introducing a supervised regularisation term into the policy objective, generalising the work of \citet{fujimoto2021minimalist}. While this helps stabilise the number of critics required across various datasets, it introduces an additional hyper-parameter needing tuning.
The performance of offline RL algorithms often relies on substantial hyper-parameter tuning across tasks and data qualities, and, as highlighted in \citet{henderson2018deep}, this can make it unclear whether such algorithms provide meaningful improvements over existing methods.

Our work builds on \citet{an2021uncertainty} and \citet{bai2022pessimistic} that demonstrate the uncertainty across an ensemble of critics can effectively penalise OOD actions. However, we leverage this idea at evaluation time rather than during training. We show that the epistemic uncertainty captured by the ensemble can indicate when the agent might propose an action at evaluation that could induce distribution shift. In such scenarios, our switching technique allows us to select actions that stay close to the data. This approach achieves the effect of stabilising the ensemble size hyper-parameter, as seen in TD3-BC-N, but in a manner that is robust to the task and data-quality.

\subsection{Offline to Online Fine-tuning}  Transitioning from offline RL to online fine-tuning presents additional challenges.  \citet{lee2022offline} note that offline algorithms are susceptible to performance drops and even collapse during the shift to online fine-tuning due to data distribution shifts. In a similar fashion to evaluation, we use the epistemic uncertainty captured by the ensemble of critics to gradually explore the environment during online training starting from regions close to the support of the data.

Methods such as TD3-BC-N introduce additional hyper-parameters to handle this transition. Furthermore, the reliance on hyper-parameters can ultimately undermine the usefulness of the theoretical guarantees that algorithms like CQL, RORL and Cal-QL \citep{nakamoto2024cal} develop in both offline and fine-tuning settings.

Our method differs by eliminating the need for additional hyper-parameters during the transition to online fine-tuning. We achieve this by using the epistemic uncertainty captured by the critic ensemble as an indicator for potential distribution shift at evaluation time. This allows us to dynamically switch between the RL policy and a BC policy, maintaining performance while adapting to new experiences. By implementing this switching mechanism at evaluation time, we aim to provide a more flexible and adaptable approach to both offline RL and the transition to online fine-tuning.

\citet{zhangpolicy} propose PEX, a method that switches between two policies with the aim of mitigating performance drops when transitioning to online fine-tuning. They achieve this by using a policy trained offline and another policy trained from scratch online. By keeping the offline policy parameters fixed during online learning they demonstrate that they can prevent collapse in performance that could occur due to the overly pessimistic critic values learnt by the offline model. However, by relying on using an uninitialised policy, at the beginning of the online fine-tuning phase, their initial performance is typically lower than can be achieved by offline RL models, diminishing one of the crucial benefits of offline RL. \citet{wang2023train} develop a framework that instead of switching between two policies, introduces another parameter that controls the policies level of pessimism in any given state. This provides a more flexible approach to controlling pessimism during online fine-tuning, but, the additional inclusion of a balance model that also needs to be trained offline can cause the method to struggle with limited data that can worsen offline and online finetuning performance as well as require longer convergence times.

Another framework that focuses solely on offline to online fine-tuning, ENOTO \citep{DBLP:conf/ijcai/ZhaoHM00M24}, uses an ensembles of critics to encourage exploration during the online fine-tuning phase. They introduce an ensembles of critics into algorithms such as CQL that they train pessimistically using the methods described by \citet{an2021uncertainty}. They then use the trained ensembles during online finetuning to encourage exploration and improve performance, using techniques such as SUNRISE \citep{lee2021sunrise} developed for the online RL setting. By taking advantage of existing methods to encourage exploration during the online fine-tuning phase, they show they can quickly improve performance in the online setting. However, by introducing an ensemble of critics offline to already pessimistic algorithms their method can also worsen offline performance.
\section{Methodology}

\subsection{Preliminaries}

We model the RL problem as a Markov decision process (MDP) that can be completely specified by the tuple $(\mathcal{S},\mathcal{A},P,\rho,\gamma)$, where $\mathcal{S}$ is the state space, $\mathcal{A}$ is the action space, $P(s'|s,a)$ defines the environment dynamics, $\rho(s_0)$ describes the initial state distribution, $r \vcentcolon \mathcal{S} \times \mathcal{A} \to \mathbb{R}$ is the reward function and $\gamma \in (0,1]$ is the discount factor. The agent's interaction with the MDP is governed by policy $\pi(a|s)$ and the expected return for a given policy can be described using the action value function $Q^\pi(s,a) = E_\pi[\sum_{t\ge0}\gamma^tr_t|s_0=s,a_0=a]$. The aim of an agent in RL is to learn an optimal policy $\pi^\star$ defined such that it maximises the expected discounted return $$\pi^\star = \arg\max_\pi J(\pi) \vcentcolon=  E_{s\sim\rho,a\sim\pi(\cdot|s)} \left[Q^\pi(s,a)\right].$$

In offline RL the agent must learn its policy by sampling experience from a fixed dataset $\mathcal{D}$ containing data collected from a behaviour policy of unknown quality. As a result of being limited to learning from a dataset, the critic values for off-policy RL methods can over-estimate actions not seen in the data, which, in turn, encourages the policy to take actions outside of the dataset resulting in a distributional shift when the policy is used for evaluation.

\subsection{Policy switching}

Policy switching is a novel strategy that combines the strengths of RL and BC to address the challenges of offline RL. The core idea is to train two policies independently and then dynamically select between them, during evaluation, based on uncertainty estimates. In RL, we aim to learn a policy that maximizes long-term rewards by considering entire trajectories of states and actions. This ability to reason about long-term consequences is a key advantage of RL over BC, which simply tries to mimic observed actions. We refer to this as "trajectory optimisation" to emphasize RL's capacity to construct potentially better action sequences than those observed in the dataset.

We denote our RL policy as $\pi_{RL}$, and define an ensemble of $N$ critics as $\{Q_i\}^N_{i=1}$ where each $Q_i$ estimates the action-value function. This ensemble plays a crucial role in estimating uncertainty, which will guide action selection during evaluation.
Concurrently, we train a BC policy, $\pi_{BC}$, using supervised learning, that aims to replicate the behaviour observed in the offline dataset. This policy serves as a fallback option when the RL policy's actions are deemed too uncertain.
During evaluation, we use uncertainty estimates derived from the ensemble of critics and the dataset itself to decide whether to use the action proposed by the RL policy or the BC policy. This dynamic selection process allows us to balance between two objectives: (a) leveraging RL's ability to optimise trajectories and potentially improve upon the behaviour in the dataset, (b) staying close to known good actions when uncertainty is high by using the BC policy. 

The intuition behind this approach is that in regions of $\mathcal{S}$ where the RL policy is confident (low uncertainty), it can likely improve upon the behaviour in the dataset through its trajectory optimisation. However, in regions where the RL policy is uncertain, it's safer to fall back on the BC policy, which closely mimics the known good behaviour from the dataset. This policy switching technique allows us to adapt to different dataset qualities without extensive hyper-parameter tuning. It also provides a natural way to transition from offline learning to online fine-tuning. In the next section, we delve deeper into how we quantify different types of uncertainty to guide our action selection process.

\subsection{Batch constrained MDP}

\citet{fujimoto2019off} show that under an MDP $\mathcal{M_B}$ with the same state and action space as the true MDP $\mathcal{M}$, but with transition probability specified by the dataset, Q-learning will converge to the optimal value function under $\mathcal{M_B}$. They also show that the extrapolation error stems from the divergence in transition distributions between $\mathcal{M_B}$ and $\mathcal{M}$. 

We hypothesise that the divergence in the transition distributions is reflected in the epistemic and aleatoric uncertainty obtained from both the dataset used and the RL model trained. By quantifying uncertainty for policy switching, we balance maximising return according to $\mathcal{M_B}$ with minimising divergence in transition distributions, enhancing offline RL performance. In the ideal scenario, with complete knowledge of uncertainty, our method would be capable of picking the best action across the two policies that would ensure it stays close to $\mathcal{M_B}$ whilst also outperforming the individual policies in the offline setting.
  
\subsection{Uncertainty quantification}

In our policy switching approach, we leverage two types of uncertainty to guide action selection: epistemic uncertainty and aleatoric uncertainty.

\subsubsection{Epistemic uncertainty} Epistemic uncertainty represents our model's uncertainty due to limited knowledge or data. In the context of our RL policy, this uncertainty is high for state-action pairs that are underrepresented in our training data. We estimate epistemic uncertainty using our ensemble of critics via the standard deviation $\sigma_Q(s, \cdot)$ defined by $$\sigma_Q^2(s, \cdot) = N^{-1} \sum_{i=1}^N (Q_i(s, \cdot) - \bar{Q} )^2.$$ A high $\sigma_Q$ indicates disagreement among critics, suggesting high epistemic uncertainty for that state-action pair. The use of variance across ensembles as a measure of epistemic uncertainty has been adopted in various areas of deep learning literature \citep{WALL2003191,Brown2020UncertaintyQI}.
We use this measure of epistemic uncertainty to define regions in the environment that are sufficiently close to the support of the data such that trajectory optimisation can be safely performed.

\subsubsection{Aleatoric uncertainty} Aleatoric uncertainty represents inherent stochasticity in the environment or dataset. In offline RL, this uncertainty is closely related to the quality and diversity of the dataset. Quantifying this uncertainty is crucial as it informs how much we can trust our RL policy to generalise beyond the observed data. Low-diversity datasets may lead to overfitting and poor performance, while diverse datasets allow for better generalisation.
Dataset diversity can be measured in various ways, from analysing state-action distributions to more complex methods involving pairwise trajectory distances or dimensionality reduction techniques. 

In this work, we propose a simple yet effective metric based on the variability of returns across trajectories. We define $\sigma_D$ as the normalised standard deviation of returns:
$$\sigma_{D} = \frac{1}{R_{max}}\sqrt{\frac{1}{N}\sum_i (R_i - \bar R)^2}, $$ where $R_i$ is the return of the $i$-th trajectory, $\bar R$ is the mean return across all trajectories, and $N$ is the number of trajectories. The division by $R_{max}$ serves as our normalization step, ensuring that $\sigma_D$ is scale-invariant and comparable across different tasks and reward scales.

This metric is motivated by the insight that diverse datasets should exhibit a wider range of outcomes. A low $\sigma_D$ indicates similar returns across trajectories, suggesting a narrow range of behaviours and necessitating heavier reliance on BC. Conversely, a high $\sigma_D$ suggests diverse behaviours, allowing our RL policy more freedom to generalise.
We choose this normalised, return-based metric for its simplicity and interpretability. While it may not capture all aspects of trajectory diversity, it provides a clear signal about dataset quality without requiring complex computations.

\subsection{Action switching mechanism}

Our policy switching mechanism aims to dynamically select between  $\pi_{RL}$ and $\pi_{BC}$ during evaluation, leveraging the uncertainty measures introduced in the previous section. The key idea is to use the RL policy when we're confident in its decisions and fall back to the BC policy when uncertainty is high. We start by defining a pessimistic Q-value estimate for the RL policy:
$$Q_{RL}(s, a_{RL}) = Q_{med}(s, a_{RL}) - f(\sigma_D) \cdot \sigma_Q(s, a_{RL})$$
where $a_{RL} \sim \pi_{RL}(\cdot|s)$ is the action proposed by the RL policy and $Q_{\text{med}}(s,\cdot)$ denotes the median Q value. Similar variants of this method for penalising Q values have been studied before in both online RL, to encourage exploration \citep{lee2021sunrise}, and offline RL \citep{bai2022pessimistic} to prevent overestimation bias during training, in contrast, we apply our formulation at evaluation instead. Our formulation is motivated by several key considerations. First, we use $Q_{med}$ instead of the mean to reduce sensitivity to outliers in the ensemble. Second, we subtract a term proportional to $\sigma_Q$ to penalise actions with high epistemic uncertainty. Lastly, we introduce a function $f(\sigma_D)$ to modulate the penalty based on the dataset's aleatoric uncertainty.

The function $f(\sigma_D)$ plays a crucial role in our policy switching mechanism by modulating the epistemic uncertainty penalty based on the dataset's aleatoric uncertainty. We require this function to satisfy several key properties. First, it should increase as $\sigma_D$ decreases, reflecting the need for greater conservatism with less diverse datasets. Second, it should be bounded to ensure the uncertainty penalty remains within a controllable range. Third, it should provide a smooth, non-linear transition between low and high penalty regimes to allow for nuanced behaviour across different uncertainty levels. To satisfy these requirements, we propose using
$$
f(\sigma_D) = m \cdot (1 - \exp(-\alpha/\sigma_D))
$$
where $m$ and $\alpha$ are hyperparameters. Here, $m$ controls the maximum value of the function, setting an upper bound on the uncertainty penalty. $\alpha$ adjusts the function's sensitivity to $\sigma_D$, particularly for small values of $\sigma_D$, influencing how quickly the penalty increases as dataset diversity decreases.

This function is well-suited for our purposes for several reasons. It is monotonically decreasing with respect to $\sigma_D$, ensuring higher penalties for datasets with lower aleatoric uncertainty (less diverse datasets). Furthermore, it is bounded between 0 and $m$, providing a controlled range for the uncertainty penalty. The exponential term creates a smooth, non-linear transition between high and low penalty regimes, allowing for fine-grained adjustment of the penalty based on the level of dataset uncertainty. By using this function, we can adaptively adjust our policy switching mechanism's conservatism based on the quality of the dataset, thereby enhancing its robustness across different offline RL scenarios.

Finally, to decide between the RL and BC policies, we compare the pessimistic Q-value of the RL action to the median Q-value of the BC action:
$$
a = \begin{cases}
a_{RL}, & \text{if } Q_{RL}(s, a_{RL}) \geq Q_{med}(s, a_{BC}) \\
a_{BC}, & \text{otherwise}
\end{cases}
$$
where $a_{BC} \sim \pi_{BC}(\cdot|s)$ is the action proposed by the BC policy.
This decision rule is motivated by careful considerations of the trade-offs between potential improvement and safety. When $Q_{RL} \geq Q_{med}(s, a_{BC})$, it suggests that even our pessimistic estimate of the RL action's value is at least as good as our estimate of the BC action's value. In this case, we trust the RL policy to potentially improve upon the dataset behaviour. Conversely, when $Q_{RL} < Q_{med}(s, a_{BC})$, it indicates that the uncertainty-adjusted value of the RL action is lower than our estimate of the BC action's value. In this scenario, we fall back to the known good behaviour captured by the BC policy.

This simple mechanism allows us to adaptively balance between exploration (trying to improve upon the dataset) and conservatism (staying close to known good behaviour) based on our confidence in the RL policy's decisions. The dynamic nature of this approach enables our algorithm to make informed choices at each step, leveraging the strengths of both policies while mitigating their respective weaknesses. We provide pseudocode for our policy switching mechanism offline in Algorithm \ref{alg:policy switching evaluation}.

\begin{algorithm}[th!]
    
   \caption{Policy switching}
   \label{alg:policy switching evaluation}
\begin{algorithmic}[1] 
    \STATE Train RL policy $\pi_{RL}$ and BC policy $\pi_{BC}$ independently using dataset
    \STATE $s_0$ = env.reset()
    \FOR{environment step $t=0,1\dots$}
    \STATE Sample RL action: $a_{RL} \sim \pi_{RL}(\cdot|s_t) $
    \STATE Sample BC action: $a_{BC} \sim \pi_{BC}(\cdot|s_t) $
    \STATE Compute uncertainty adjusted critic value for reinforcement learning action:
        $$Q_{RL} = Q_{\text{med}}(s,a_{RL}) - f(\sigma_{D})*\sigma_Q(s,a_{RL})$$
    \STATE Compute average value estimate for BC action 
        $$Q_{BC} = Q_{\text{med}}(s,a_{BC}) $$
    \IF{$Q_{RL} \ge Q_{BC}$}
    \STATE $a = a_{RL}$
    \ELSE
    \STATE $a = a_{BC}$
    \ENDIF
    \STATE Act in environment: $r_t, s_{t+1} \leftarrow \text{env.step}(a)$
    \ENDFOR
\end{algorithmic}
\end{algorithm}

\subsection{Offline to online fine-tuning}\label{sec:finetuning}

After an initial policy is derived from the static dataset, it may be necessary to fine-tune this policy in an online environment to address potential issues of distributional shift and overfitting. During online fine-tuning, the agent interacts with the environment to collect new data, allowing it to refine its policy by adjusting to discrepancies between the dataset's distribution and actual environment dynamics. This process helps mitigate the risk of poor generalisation and enhances the policy's robustness and adaptability, leading to improved performance in real-world scenarios.

Our policy switching method seamlessly extends to online fine-tuning. During this phase, we maintain the action switching mechanism but only update $\pi_{RL}$ and its critic ensemble while keeping the BC policy, $\pi_{BC}$, fixed. This approach allows us to adapt to new experiences while retaining a stable anchor to the offline dataset. 

We gradually replace offline samples with new experience online. This ensures a smooth transition without requiring an explicit mixing ratio hyperparameter. To prevent over-pessimism during critic updates, we adopt a strategy inspired by REDQ \cite{chen2021randomized}, where we randomly select a subset of two critics for each update (see Appendix for further details). As more data is collected, the epistemic uncertainty in new regions of the state space decreases, leading to an automatic annealing of BC action usage over time. This approach induces a gradual change in distribution, resulting in stable fine-tuning performance without additional modifications or hyper-parameters.

\begin{table*}[th!]
\begin{center}
\begin{small}
\caption{Normalised average returns on D4RL tasks. All algorithms are evaluated after 1M and the evaluation score is averaged across 5 seeds with 10 episodes per seed. Ensemble-based methods use $N = 10$. In brackets we provide the rank of an algorithm for each dataset as well as a final ranking based on the average rank across all datasets in the final row. We report the mean $\pm$ 95\% confidence intervals for the average return over 10 episodes for TD3-N and our method.}
\label{table:offline mujoco results}
\begin{tabular}{l|cccccccc|cr}
\toprule
Dataset & BC & IQL & CQL & Cal-QL & ENOTO & FamO2O & TD3-BC-N & TD3-N & TD3-N+BC (PS) \\
\midrule
halfcheetah-med-rep   &  36.3 (9) & 44.2 (8)& 45.0 (6) & 45.8 (4)& 45.4 (5) & 44.3 (7)& 54.7 (3)& \textbf{62.1 $\pm$ 0.76} (1)& \textbf{61.4 $\pm$ 0.58}  (2)\\
hopper-med-rep &   26.6 (9) & 91.2 (6)&  95.1 (5)& \textbf{99.6} (2) & 47.3 (8)& 87.6 (7) & 99.1 (4)& 99.3  $\pm$ 1.21 (3)& \textbf{99.7 $\pm$  0.72} (1)\\ 
walker2d-med-rep   & 23.48 (8) & 87.8 (3)&  73.1 (6) & 86.3 (5)& -0.14 (9) & 61.6 (7)& \textbf{92.4} (1) & 87.9 $\pm$ 2.35 (2) & 86.9 $\pm$ 2.17 (4)\\
\bottomrule
halfcheetah-med   & 41.9 (9)  & 48.6 (4)& 47.0 (8)& 48.0 (5)& 48.0 (5)& 47.6 (7)&64.2 (2)& \textbf{65.0 $\pm$ 0.55} (1)& 62.9 $\pm$    0.5  (3)\\
hopper-med     & 51.5  (9)  & 61.0 (6)&  59.1 (8)& 62.0 (5)& 59.7 (7)& 63.0 (4)& \textbf{101.5} (2)& 100.7 $\pm$ 7.76 (3)& \textbf{101.8 $\pm$ 4.05} (1)\\
walker2d-med   &  71.8 (8) & 73.8 (7)&  80.8 (6)& 83.3 (4)& 68.4 (9)& 82.0 (5)& 91.9 (3)& \textbf{101.3 $\pm$  2.35} (1) &98.2 $\pm$ 1.33 (2)\\
\bottomrule
halfcheetah-med-exp    & 60.1 (9) & 94.9 (3)&  95.6 (2)& 63.9 (8)& 85.4 (6)& 83.4 (7)& \textbf{100.6} (1) & 94.2 $\pm$ 5.29 (4)&93.6 $\pm$  8.39 (5)\\
hopper-med-exp   &  51.3 (9) & 97.9 (5) &  99.3 (3)& \textbf{112.5} (1)& 98.3 (4)& 81.9 (7)& \textbf{112.3} (2) &  61.5 $\pm$  29.2 (8) &88.6  $\pm$  23.8 (6)\\
walker2d-med-exp  &  107.7 (8) & 111.6 (4)&  109.6 (6)&  109.3 (7)& 92.1 (9)& 109.8 (5)&  113.9 (3) & \textbf{116.8 $\pm$  1.54} (1) & \textbf{115.9 $\pm$ 0.77} (2)\\
\bottomrule
halfcheetah-exp     &   92.6 (6) & 89.8 (7) &  - & 95.7 (2)& 93.4 (5)& 94.0 (4)& \textbf{100.4} (1) &  75.5  $\pm$ 14.4 (8) &94.7 $\pm$  5.63   (3) \\
hopper-exp   &  110.6 (3)  & 99.5 (6) &   - & \textbf{112.3} (1)& 105.0 (5)& 106.0 (4) & \textbf{112.2} (2)&  5.3 $\pm$  4.69 (8) & 37.8  $\pm$ 34.1 (7)\\
walker2d-exp  &   108.2 (6) & 112.7 (2) &  -  & 109.8 (4)& 110.1 (3)& 109.8 (4)& \textbf{113.2} (1)&  26.7 $\pm$ 13.3 (8)& 105.4 $\pm$ 7.78 (7)\\
\toprule
\textbf{Average rank} & 7.75 (9) & 5.08 (5) & 5.56 (6)  & 4 (3)  & 6.25 (8) & 5.67 (7) & \textbf{2.08} (1) & 4 (3) & 3.58 (2) \\   
\bottomrule
\end{tabular}
\end{small}
\end{center}

\end{table*}

\section{Experimental results}

In this section, we investigate the performance and stability of policy switching. For all experiments, we use Gaussian BC and TD3-N, an ensemble variant of TD3 \citep{fujimoto2018addressing}, as our base BC and RL agent respectively. We compare our method against the same RL agent without policy switching as well as several state-of-the-art algorithms for both offline learning and offline to online fine-tuning.

\begin{figure*}[th!]
\centering
\includegraphics[width=0.95\linewidth]{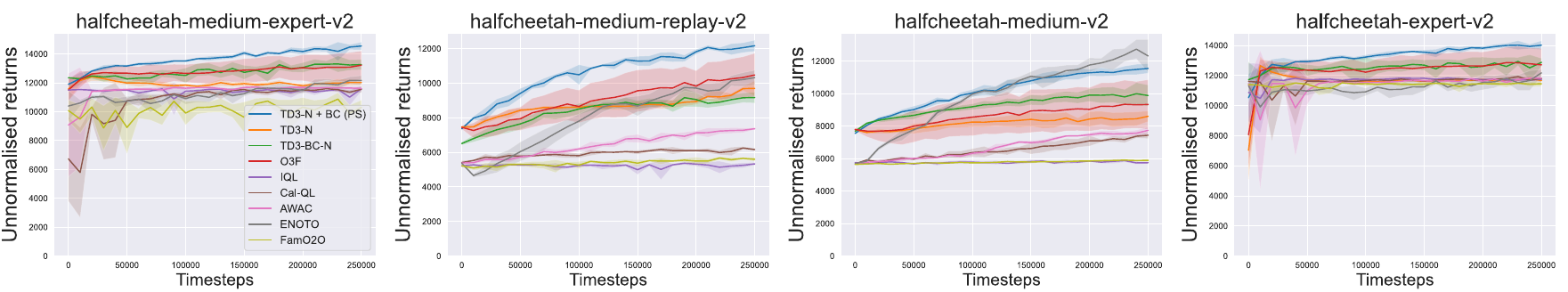} \\
\includegraphics[width=0.95\linewidth]{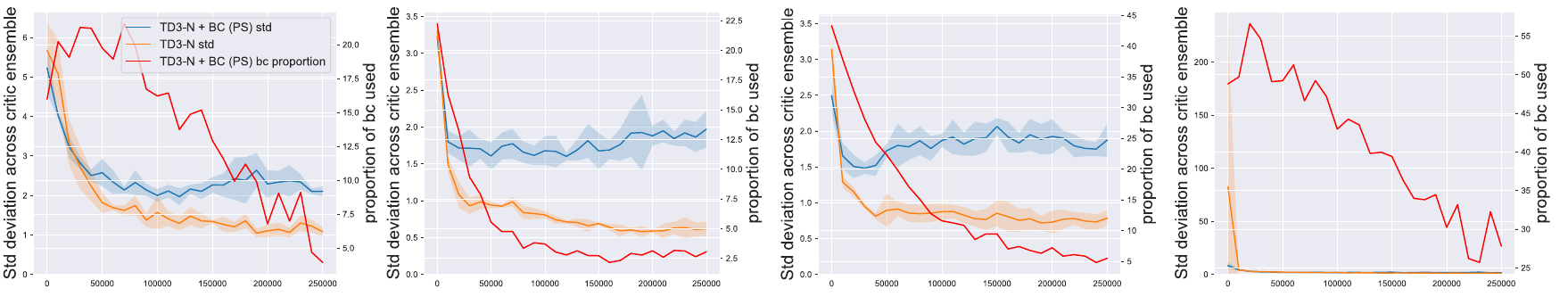}
\caption{
The top row shows the learning curves for online fine-tuning on the MuJoCo datasets. Algorithms are evaluated every 10k steps and averaged over 5 seeds and 10 episodes per seed. Bottom row shows how standard deviation across the ensemble of critic evolves for our algorithm in contrast to TD-N (blue and yellow line y-axis LHS) as well as the average proportion of BC used across an episode for our algorithm (red line y-axis RHS).
}
\label{fig:halfcheetah online results}
\Description{Plots of the unnormalised returns for all halfcheetah tasks as well as plots of how the ensemble uncertainty evolves during online fine-tuning for the corresponding tasks.}
\end{figure*}

\subsection{Offline RL tasks and datasets}

For our evaluations, we use a range of tasks from the D4RL suite  \citep{fu2020d4rl}, which provides diverse environments and datasets to assess the robustness and generalisation of our approach, and is regularly used as benchmarks for offline RL algorithms. These include the MuJoCo tasks that are composed of several locomotion problems within the continuous control setting. In these problems, rewards are provided according to the agents ability to learn precise control and balance of a robotic model whilst in motion. Datasets for these tasks are generated using a policy trained online and the quality of the dataset is determined by the quality of the policy used to collect the data. The different quality data allows us to investigate how well we can stitch together behaviour from the dataset to improve upon the policy used to generate the dataset. 

We also investigate how our method works on the more challenging AntMaze problems. Similar to the MuJoCo setting, the AntMaze tasks challenge the agent to learn how to control the navigation of ant quadruped robot but with the aim of reaching a desired location within a maze. Unlike the MuJoCo setting, rewards are only provided upon successfully reaching the desired goal. Sparse reward problems are notoriously difficult for RL algorithms both in the online and offline setting as there is very little reward that can be back-propagated to affect behaviour which typically results in a slower learning process and worse sample efficiency.

\subsubsection{Comparisons and baselines} We compare our method against algorithms developed for both offline RL and offline to online fine-tuning, including TD3-N, the base RL algorithm we use for policy switching. TD3-N is an extension of TD3 \citep{fujimoto2019off}, inspired by \citet{an2021uncertainty}, that uses an ensemble of $N$ critics to induce more pessimistic Q-values. For our experiments, we fix the size of the ensemble as $N=10$. We also compare our method to offline algorithms such as CQL \citep{kumar2020conservative}, IQL \citep{kostrikov2021offline} and TD3-BC-N \citep{beeson2024balancing}. For online fine-tuning, we also include algorithms that specifically focus on this, such as AWAC \citep{nair2020awac}, O3F \citep{mark2022fine}, Cal-QL \citep{nakamoto2024cal}, ENOTO \citep{DBLP:conf/ijcai/ZhaoHM00M24} and FamO2O \citep{wang2023train}. ENOTO uses different algorithms for evaluating performance on AntMaze and MuJoCo, for consistency we stick to using the algorithm tested on the MuJoCo datasets.

\subsubsection{Online fine-tuning}  Using the simple strategy outlined in Section \ref{sec:finetuning}, we empirically demonstrate that our policy switching technique induces a gradual distribution shift, resulting in stable fine-tuning performance. Notably, we achieve this without further modifications to our algorithm or the introduction of additional hyper-parameters. This is in contrast to methods such as TD3-BC-N and Cal-QL, which require specific adaptations for the online phase.

\begin{table*}[th!]

\begin{center}
\begin{small}

\caption{Normalised average returns on D4RL AntMaze tasks, 100 evaluations per seed over 3 seeds. Where ensembles are used, ensemble size $N = 10$. In brackets we provide the rank of an algorithm for each dataset as well as a final ranking based on the average rank across all datasets in the final row. We report the mean $\pm$ 95\% confidence intervals for the average return over 100 episodes for TD3-N and our method.}
\label{table:offline antmaze results}
\begin{tabular}{l|ccccccc|cr}
\toprule
 & BC & IQL & CQL & Cal-QL & FamO2O & TD3-BC-N & TD3-N & TD3-N+BC (PS) \\
\midrule
antmaze-umaze &  86 (5) &   83  (6) & 92.8  (2) &   72.3 (7) & 87 (4)&  91.3  (3) & 37.7  $\pm$ 9.76 (8)& \textbf{93.7 $\pm$ 2.85} (1)\\
antmaze-umaze-diverse & 87.3 (3)&  63  (4) &  37.3 (5)&  5 (6) & 69.7 (7)&  91 (2) & 0 (7)& \textbf{94 $\pm$ 4.1} (1)\\
antmaze-medium-play &   0 (8)&  69 (4) &    65.8 (5) &  73 (3)&  74.7 (2) & \textbf{80.7} (1) &  8.3 $\pm$ 12.5 (7) &  63.3 $\pm$ 1.73 (6)\\ 
antmaze-medium-diverse &  0 (8)&   70  (3)&    67.3 (5) &  67.7 (4)&  56.3 (6)& \textbf{87} (1) &  2.7 $\pm$ 2.36 (7)& 71 $\pm$ 9.26 (2)\\
antmaze-large-play &  0 (7)&  38 (4) & 20.8  (6)   &   31.7 (5)&  42 (2)& \textbf{76.7} (1)&  0 (7)&  41 $\pm$ 17.7 (3)\\
antmaze-large-diverse&    0 (7) &   38.7 (4) &   20.5 (5) &   14 (6)&  53.3 (2) & \textbf{71} (1) & 0 (7) & 49.7 $\pm$ 10.8 (3)\\
\toprule
\textbf{Average rank} & 6.33 (7)& 4.17 (4) & 4.67 (5) & 5.17 (6) & 3.83 (3) & \textbf{1.5} (1) & 7.17 (8) & 2.67 (2) \\
\bottomrule
\end{tabular}
\end{small}
\end{center}

\end{table*}

\subsection{Locomotion tasks}

\subsubsection{Offline RL} We first assess our method on the MuJoCo datasets. We train  both the RL and BC agents independently which allows us to train them concurrently, mitigating additional computational overhead costs. We evaluate our algorithm across different quality datasets and compare our policy switching technique with the individual algorithms used in Table \ref{table:offline mujoco results}. We emphasize that in producing the results for our method, we do not adjust any hyper-parameters for any task or dataset quality.

Table \ref{table:offline antmaze results} shows that our policy switching algorithm  is able to closely match, and in some cases exceed the score of the base RL algorithm in the medium-replay datasets where our measure of aleatoric uncertainty is higher. Not only are we able to produce competitive scores, but in comparison to the results from TD3-N, we are able to reduce variance for most tasks induced as a result of using different seeds.

The diversity of data in the medium-replay datasets means unconstrained off-policy algorithms are less prone to over-estimation bias and are therefore more able to accurately generalise and use trajectory optimisation to improve behaviour. In fact, the scores obtained from TD3-N on the medium replay datasets are competitive with the current best offline algorithms. In contrast, TD3-N struggles substantially on the expert datasets due to poor generalisation on OOD actions. By switching to actions from the BC agent we are able to consistently outperform TD3-N on all expert datasets. In fact, for walker2d-expert we are able to approach the BC score, which is competitive with SOTA algorithms, and in halfcheetah-expert we outperform both BC and TD3-N, indicating our methods effectiveness at balancing trajectory optimisation whilst staying close to the data. 

In the hopper-expert and medium-expert dataset, although we improve on the performance of TD3-N, we still struggle to achieve a competitive score. We observe during training that the narrowness of the data results in overconfidence from the ensemble of critics that causes poor generalisation of uncertainty outside the support of the data. This prevents the ensemble from being as useful a measure of epistemic uncertainty compared to in other datasets. Methods such as EDAC \citep{an2021uncertainty} that encourage diversity across the ensemble of critics can help to mitigate this issue but at the expense of performance on other datasets and additional computational costs.

Furthermore, by ranking the algorithms across all datasets, we can clearly see that our method gives rise to strong offline performance that other methods compromise by focusing solely on online finetuning performance.  
\subsubsection{Online fine-tuning}

Figure \ref{fig:halfcheetah online results} shows that our method is able to consistently exceed the benchmarks for offline to online fine-tuning across all dataset qualities with no hyper-parameter tuning required. We also see how the proportion of behavioural cloning being used is naturally annealed as the variance across the critic ensemble drops with the influx of online data, which we observe helps mitigate the risk of an initial drop in performance.

\subsection{AntMaze tasks} 

\subsubsection{Offline RL} 
For the AntMaze tasks we compare offline performances in Table \ref{table:offline antmaze results}. Both the TD3-N and BC agent are unable to successfully learn good behaviour using datasets from all but the smallest sized maze on their own. Despite this, by using the epistemic uncertainty of the critic ensembles we are able to stitch together behaviour from these two algorithms to obtain performance that far exceeds both algorithms on all datasets.

In the MuJoCo tasks we were able to use the variance of returns across trajectories to measure aleatoric uncertainty which was possible due to the tasks using dense rewards. The AntMaze tasks, in contrast, only provide rewards at the end of a trajectory which make them not as effective for measuring the aleatoric uncertainty. To keep with the idea of simple design choices, we use the variance across trajectory lengths in the AntMaze tasks as a measure of aleatoric uncertainty across datasets. We re-adjust the hyper-parameter $\alpha$ for using this new measure for uncertainty and then keep it fixed for all tasks and datasets. 

Although we are able to achieve results competitive with SOTA algorithms for a few datasets we note a bigger drop in performance compared to TD3-BC-N. This is in part due to the fact their algorithm is tuned for each maze type, however, we also highlight that as the length of the trajectory increases there is an increased probability of selecting an action leading to a region outside of the support of the data. This probability is linked to how well we can determine whether BC or RL should be selected. Despite this, we are still able to outrank all other offline algorithms.

\subsubsection{Online fine-tuning}
\begin{figure}[th!]
    \centering
    \includegraphics[width=0.95\linewidth]{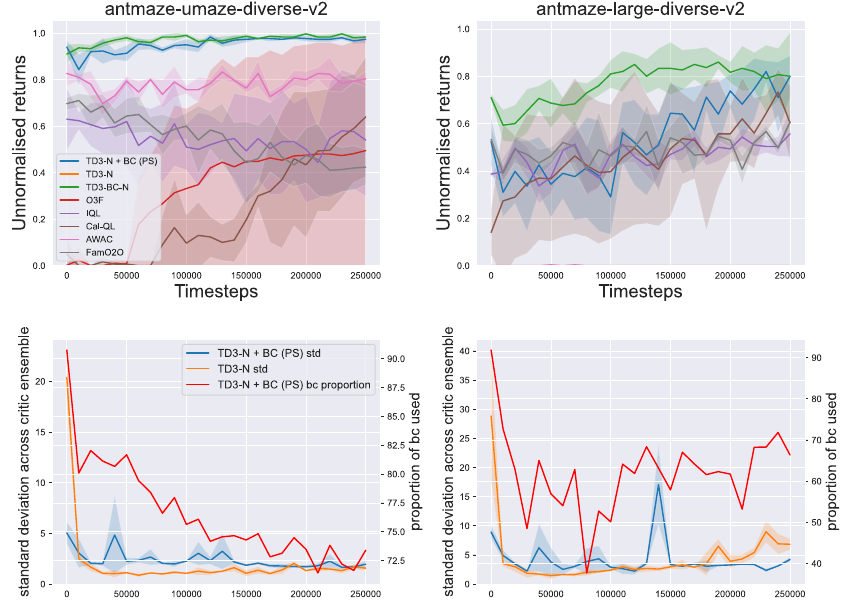}
    \caption{The top row shows fine-tuning curves on the AntMaze datasets. Algorithms are evaluated every 10k steps averaged over 3 seeds and 100 episodes per seed. Bottom row shows evolution of standard deviation across the ensemble of critics and the average proportion of BC.}
    \Description{Plots of the unnormalised returns for antmaze umaze diverse and antmaze large diverse tasks as well as plots of how the ensemble uncertainty evolves during online fine-tuning for the corresponding tasks.}
    \label{fig:antmaze online results}
\end{figure}

We present our results in Figure \ref{fig:antmaze online results}. In the umaze-diverse environment we see that despite our algorithm starting with a high offline RL score there is little to no dip in initial performance. In contrast, TD3-N's performance collapses in the umaze-diverse environment and is unable to recover during the remainder of online-tuning. We also note that Cal-QL performs well on the large Antmaze datasets but struggles on umaze-diverse and isn't able to reach the same performance as our policy switching method on umaze. 

The bottom row of Figure \ref{fig:antmaze online results} shows a similar trend to the MuJoCo fine-tuning results, with the proportion of BC used gradually being annealed as the critic ensemble variance drops. Unlike the results in Figure \ref{fig:halfcheetah online results}, the amount of BC looks to converge to a non-zero value even for datasets umaze and umaze-diverse where our algorithm is able to reach a near perfect performance.
We believe this dynamic balance between ensemble variance and amount of BC used is key to preventing a collapse in learning as observed with TD3-N.

\section{Ablation studies}

We investigate the effect our measure of aleatoric uncertainty has on the robustness of the policy switching mechanism by comparing it to one where $f(\sigma_D)$ is fixed across a selection of tasks and datasets in Table \ref{tab:ablation aleatoric}. For the fixed setting, as $f$ is bounded between $[0,m)$, we fix it at $m/2$. The results indicate that factoring in uncertainty from the dataset results in a clear improvement across the selected tasks.

\begin{table}[th!]
    \centering
    \caption{Effect the aleatoric uncertainty, $\sigma_D$, on the offline performance of our policy switching mechanism. Results are averaged over 5 seeds.}
    \begin{tabular}{l|c|c}
        \toprule
         &  PS w/ fixed $\sigma_D$  & PS \\
         \midrule
         halfcheetah-exp  & 83.1 $\pm$ 19.1 & \textbf{94.1 $\pm$ 6.4}  \\
         halfcheetah-med-rep  & 60.12 $\pm$ 0.61 & \textbf{61.4 $\pm$ 0.66} \\
         walker2d-med & 96.8 $\pm$ 1.6 & \textbf{98.2 $\pm$ 1.52}\\
         walker2d-exp & 80.9 $\pm$ 18.53 & \textbf{105. 4 $\pm$ 8.87} \\
         hopper-med & \textbf{102.7 $\pm$ 0.68} & 101.8 $\pm$ 4.63 \\
         hopper-exp  & 25 $\pm$ 25.3 &  \textbf{37.8 $\pm$ 38.9} \\
         \bottomrule
    \end{tabular}
    \label{tab:ablation aleatoric}
\end{table}

We also provide additional experiments exploring the effects of adjusting the hyperparameter $\alpha$, as well as a comparison between using a vanilla BC agent and Gaussian BC agent within our policy switching mechanism in the appendix.

\section{Discussion and conclusions}

\balance

In this work, we developed a policy switching technique as an alternative way to combine RL and BC algorithms to learn behaviour from offline data. Empirically, we demonstrate that our method rivals state-of-the-art algorithms in both offline and offline-to-online fine-tuning settings, achieving this with minimal hyper-parameter tuning and algorithm modification. Our method is simple to implement, requiring only minor modifications to existing baseline algorithms and incurring minimal computational overhead as both algorithms are trained independently and can be run concurrently.

Future work could explore alternative methods for quantifying epistemic uncertainty, potentially enabling the incorporation of algorithms like IQL \citep{kostrikov2021offline}. An interesting avenue for theoretical investigation emerges from the parallels between our method and the concepts introduced by \citet{kumar2022should}. They define critical and non-critical states based on the impact of actions on overall RL agent performance. Their offline RL algorithm achieves better performance guarantees than BC by selecting improved actions in non-critical states while matching BC performance in critical states. This is achieved by updating the policy and critic only with state-action pairs close to the data, similar to IQL. Our method can be viewed as treating states with high epistemic uncertainty as critical states, analogous to their approach. However, unlike their method which requires approximating environment dynamics, our switching approach potentially makes us less reliant on data quality and environment complexity. Future theoretical work could formalise these connections, potentially leading to stronger performance guarantees for our policy switching method in both offline and online settings.


\begin{acks}
    This work was supported by the Engineering and Physical Sciences Research Council (2674942) and a UKRI Turing AI Acceleration Fellowship (EP/V024868/1).
\end{acks}



\bibliographystyle{ACM-Reference-Format} 
\bibliography{main}


\end{document}